\documentclass[letterpaper, 10pt, conference]{ieeeconf}
\IEEEoverridecommandlockouts 
\usepackage{graphics}
\usepackage{epsfig}
\usepackage{amsmath}
\usepackage{xcolor}
\usepackage{bm}
\usepackage{array}
\usepackage[super]{nth}
\usepackage{siunitx}
\usepackage{upgreek}
\usepackage{amssymb}
\usepackage[hidelinks]{hyperref}
\usepackage{cite}

\title{\LARGE \bf
The ConScenD Dataset: Concrete Scenarios from the highD Dataset According to ALKS Regulation UNECE R157 in OpenX}

\author{Alexander Tenbrock$^{1}$, Alexander König$^{1}$, Thomas Keutgens$^{1}$, Julian Bock$^{1}$,\\ Hendrik Weber$^{2}$, Robert Krajewski$^{2}$ and Adrian Zlocki$^{1}$
	\thanks{$^{1}$The authors are with the Automated Driving Department of fka GmbH, 52074 Aachen, Germany
		{\tt\small \{alexander.tenbrock, alexander.koenig, thomas.keutgens, julian.bock, adrian.zlocki\}@fka.de}}%
	\thanks{$^{2}$The authors are with the Institute for Automotive Engineering at RWTH Aachen University 
		{\tt\small \{hendrik.weber, robert.krajewski\}@ika.rwth-aachen.de}}}%

\begin{document}
	\maketitle
	\thispagestyle{empty}
	\pagestyle{empty}


\begin{abstract}
With Regulation UNECE R157 on Automated Lane-Keeping Systems, the first framework for the introduction of passenger cars with Level 3 systems has become available in 2020.
In accordance with recent research projects including academia and the automotive industry, the Regulation utilizes scenario based testing for the safety assessment. The complexity of safety validation of automated driving systems necessitates system-level simulations.
The Regulation, however, is missing the required parameterization necessary for test case generation.
To overcome this problem, we incorporate the exposure and consider the heterogeneous behavior of the traffic participants by extracting concrete scenarios according to Regulation's scenario definition from the established naturalistic highway dataset highD.
We present a methodology to find the scenarios in real-world data, extract the parameters for modeling the scenarios and transfer them to simulation.
In this process, more than 340 scenarios were extracted. 
OpenSCENARIO files were generated to enable an exemplary transfer of the scenarios to CARLA and esmini.
We compare the trajectories to examine the similarity of the scenarios in the simulation to the recorded scenarios.
In order to foster research, we publish the resulting dataset called ConScenD together with instructions for usage with both simulation tools.
The dataset is available online at https://www.levelXdata.com/scenarios.

\end{abstract}

\section{Introduction}

Implementations of automated driving functions in prototypes and test drives have already been demonstrated in the past. Still, despite several announcements by the automotive industry, Level 3 automated driving systems (L3-ADS) \cite{J3016} or higher have not yet been introduced to the market. One of the reasons is the long-standing lack of a legal and regulatory basis for safety assessment of these systems.

For the introduction of L3-ADS, there must be proof of a positive impact on traffic safety, resulting in a tremendous testing effort \cite{winner2015quo, LindmannTestKm}. With an increasing level of automation, the driver’s responsibilities decrease and the complexity of the automated driving function grows. Complementing prototype-based real-world tests with virtual tests helps to diminish those efforts, though without replacing prototype tests entirely \cite{VirtuelleTests}.

Several research projects like PEGASUS \cite{PegasusOverview}, ENABLE-S3 \cite{EnableS3Overview}, SAKURA \cite{SAKURA} or CATAPULT \cite{CATAPULT} have developed frameworks for the safety assessment process of these vehicles based on traffic scenarios. The central element of their frameworks is a database combining different data sources and a processing chain that generates test specifications. Through a variety of data sources like accident databases, field operational tests (FOT) or driving simulator studies, relevant scenarios are extracted and fed into the database. Those scenarios are used, among other things, to generate test cases either for simulated environments or real-world test tracks.

The UNECE Regulation \textit{Proposal for a new UN Regulation on uniform provisions concerning the approval of vehicles with regards to Automated Lane Keeping System} \cite{ALKS} provides the first regulatory framework for the introduction of L3-ADS into the market in Europe. The Regulation provides a minimum set of scenarios for safety validation. A technical service is supposed to select and vary the concrete parameters of these test scenarios in a reasonable manner. This is often linked to the risk of a test scenario based on concrete parameters \cite{thorn2018framework}. As risk is typically dependent on the probability of exposure and potential severity of a scenario, it is an important step to determine the exposure for scenarios \cite{ISO26262}. The exposure is used to quantify the probability of occurrence for certain characteristics of a scenario.

A valid methodology to get the exposure is the analysis of real-world data, which has not been conducted for the ALKS scenarios. Thus, this paper contributes an approach to extract concrete scenarios, i.e. concrete parameter sets of a scenario, from measured trajectory datasets. From this, a dataset of concrete scenarios is created and made publicly available. The presented methodology intends to enrich the Automated Lane-Keeping System (ALKS) scenarios with information on realistic scenario parameters derived from real-world data. As it offers recordings of naturalistic traffic behavior, highD \cite{highDdataset}, a large-scale trajectory dataset from German Autobahn by fka and ika, is utilized to get the exposure on the behavior of the traffic participants. The concrete scenarios are provided in the OpenX standards OpenSCENARIO and OpenDRIVE for flexible usage in diverse simulation tools.

This paper is structured in the following way: Section II gives a short introduction into relevant related work regarding the use of scenarios during the safety assessment process, the Regulation on L3-ADS systems, scenario simulation standards and finally the used dataset. Section III highlights our method for the extraction of concrete scenarios from real-world data and its transformation process to the simulation using ASAM OpenX. Section IV describes the extracted scenarios as they are transferred into the simulation and presents the differences between the real-world scenarios and their counterparts in the simulation. Finally, section V gives a conclusion.

\section{Related Work}

\subsection{Scenario Definition}

Due to the extensive use of scenario based testing, it is important to have a common understanding of the term "scenario". In general, a scenario refers to the abstraction and general description of a temporal and spatial traffic constellation \cite{SAE91381}. Since an easy definition is not possible, several contributions have extended the definition. 

In \cite{bagschik2017szenarien} three abstraction levels called functional, logical and concrete have been established to describe a scenario in terms of different level of detail. Functional scenarios can either be described with an informal way of description like pictograms, free text or based on a predefined machine-readable scheme. The transfer from the linguistic description to a state-space description enables the derivation of technical requirements including valid and non-valid value ranges (logical scenarios). Finally, concrete scenarios not only represent the lowest level of abstraction with the highest variety of scenarios but also shape executable test cases. In concrete scenarios an explicit assignment of parameters from the state space to a concrete scenario unambiguously describes all entities and their relationships. Depending on the discretization of parameters, a logical scenario can be the origin for an infinite number of concrete scenarios. The outcome of the scenario is, as already in the case of the logical scenario, hypothetical and not further defined since the behavior of the ego vehicle is not specified \cite{SAE91381}. Therefore, concrete scenarios in particular are used to generate test cases and test specifications.

\subsection{Gathering of Concrete Scenarios}
Two main approaches are used to identify scenarios: knowledge-based and data-driven. While in the former experts define scenarios from top-down the latter identifies scenarios as a result of clustering measurement data. However, neither of the approaches excludes experts’ knowledge or support of data and measurements \cite{riedmaier2020survey}.

Knowledge-based approaches implement a knowledge representation, e.g. in the form of an ontology, and generate a scenario catalog based on the ontology \cite{bagschik2018wissensbasierte}. The completeness of the scenario catalog can be considered using the ontology, which can generally be implemented more efficiently. Though, the derived scenarios could lack real-life representation.

Data-driven approaches leverage the iterative improvement by collecting new data to create extensive and plausible scenario catalogs. Nevertheless, the approach could perform poorly in covering high diversity if a bad database is used. While most of the data-driven approaches identify logical scenarios based on a parameterizable model \cite{weber2019framework}, some are utilizing machine-learning methods \cite{krajewski2019beziervae, wang2020clustering}. For this purpose, the method of unsupervised machine learning is used, which divides a data set into scenarios without previously defined knowledge. When using machine learning, the completeness of the scenario catalog depends on the traffic situations contained in the data set. Furthermore, an additional effort is required to assign a meaning to the individual scenarios, for example, to evaluate the relevance of the scenarios for the safety assessment.

\subsection{OpenSCENARIO \& OpenDRIVE}

For the unambiguous description of scenarios and the efficient integration into different simulation environments, standards are developed for their description. With OpenSCENARIO and OpenDRIVE two standards for the harmonization of simulation interfaces in the field of automotive simulation are being pushed by the Association for Standardization of Automation and Measuring Systems (ASAM) \cite{asamOD, asamOS}. Since both standards are used in this research, they are briefly introduced. OpenDRIVE allows the description of a static environment including e.g. roads and signs. This enables that individual road elements such as straight lines and curves can be parameterized and combined with other elements to form entire routes. 

Unlike OpenDRIVE, OpenSCENARIO is used to describe the temporally dynamic elements of a scenario. For example, the position of a traffic light is defined in OpenDRIVE, while the switching times are defined in OpenSCENARIO. The main focus is on the road users. Vehicles can be defined in catalogs and positioned on a route that is referenced in OpenDRIVE. The behavior of vehicles in the scenario can be described by a storyboard, which includes actions that can be triggered when predefined conditions are met. The simulative use case does not directly translate to an evidence-based use case. For example, actors in the simulation often perform maneuvers when certain conditions are met. Examples of such triggering conditions (triggers) are:
\begin{itemize}
    \item Position in longitudinal or lateral direction of a traffic participant
    \item Distance between two traffic participants
    \item Absolute and relative velocity
    \item Time gap, Time-to-Collision
\end{itemize}

If a vehicle is given a new target lane, then the lateral motion can be described by geometric primitives or precisely specified by a trajectory in the form of a list of positions.

\subsection{Simulation Tools and Interfaces}

For simulating OpenDRIVE and OpenSCENARIO files a multitude of simulation environments with different strengths and capabilities exist. Among them are several commercial tools like VTD \cite{VTD}, dSPACE ASM \cite{dSpaceASM} and multiple open-source tools like esmini \cite{esmini}, openPASS \cite{dobberstein2017eclipse}, and Carla \cite{dosovitskiy2017carla}.
In order to enable a wide accessibility and applicability of the generated database, the two open-source tools esmini and CARLA were picked for testing the generated scenario files.

Esmini is a basic and lightweight OpenSCENARIO player. It is the result of the Swedish research project Simulation Scenarios \cite{SimS} and has the advantage of a very good OpenSCENARIO coverage, offers however only basic 3D visualization capabilities. The 3D view is based on OpenSceneGraph, an open-source graphics library that focuses more on stability and performance than on complex and accurate 3D visualizations.  

CARLA is an open-source driving simulator for autonomous driving research \cite{dosovitskiy2017carla}. It serves multiple purposes like learning driving policies, training perception algorithms, etc. It is based on Unreal Engine and thus enables a much more complex graphical representation of the environment than esmini. An extension available for CARLA is the ScenarioRunner \cite{CarlaSR}, which is a module allowing the definition, simulation, and evaluation of complex traffic scenarios. 

It is important to mention that, at the time of writing, the various tools do not support all the attributes of the OpenSCENARIO standard. Therefore specific adaptations have to be made for the tool in question.

\subsection{Regulation on ALKS}
With the adoption of the UN Regulations on the requirements for ALKS \cite{ALKS} by UNECE's World Forum for Harmonization of Vehicle Regulations, the first binding international regulatory basis on Level 3 automation has been set. The intention of the Regulation is to establish uniform provisions concerning the approval of vehicles with regard to ALKS and its administrative processes suitable for e.g. type approval, technical requirements, audit and testing provisions.

ALKS controls the lateral and longitudinal movement of the vehicle for extended periods without further driver command meaning that the activated system is in primary control. The system can only be activated under certain conditions on roads including segments, where pedestrians and cyclists are prohibited and which are equipped with a physical separation. The application is limited to an operational speed of up to 60 km/h maximum and passenger cars \cite{ALKS}.

In order to verify the technical requirements of the ALKS system, the Regulation defines tests outlined in Annex 5 at a functional level, which are to be carried out in close coordination with a technical service, who is in charge for selecting the specific test parameters. According to the document, the manufacturer shall declare the system boundaries and define different combinations of test parameters (e.g. speed, type and offset of target, curvature of lane) in order to cover scenarios in which a collision shall be avoided by the system as well as those in which a collision is not expected to be avoided, where applicable. However, the parameter selection remains to the manufacturer technical service, which have to find a reasonable parameterization that represent real-world traffic. Additionally, the Regulation also states requirements on pass-fail-criteria that allow the derivation not only for a given set of test parameters, but for any combination of parameters in which the system is designed to work.

The test scenarios are foreseen to assess the performance of the system with regard to the dynamic driving task regarding six different groups of tests. The groups are focusing on aspects like lane keeping, object blocking the lane, follow lead vehicle, cut-in, cut-out and field of view. Tests involving avoidable as well as unavoidable collisions must be conducted to examine the behavior of an ALKS under test.

In \cite{BmwALKS} test scenarios derived from the ALKS Regulation have been implemented in a bundle of XML files according to the standards OpenSCENARIO and OpenDRIVE, which are executable with standard compliant simulators. Since the Regulation leaves room for interpretation, the coordination of a common interpretation has been pushed with this work. The publication provides 15 executable test scenarios derived from the six subject areas analogous to Annex 5, Chapter 4.1-4.6 as an initial attempt to clarify the described set of functional scenarios. For each of the 15 test cases one specific parameter set has been uploaded. Due to the heterogeneous behavior of the traffic participants, a greater number of scenarios for each test case must be implemented to test the ALKS system and to account for the exposure of these scenarios.

\begin{figure}[t]
        \vspace*{0.2cm}
		\centering
		\includegraphics[width=0.48\textwidth]{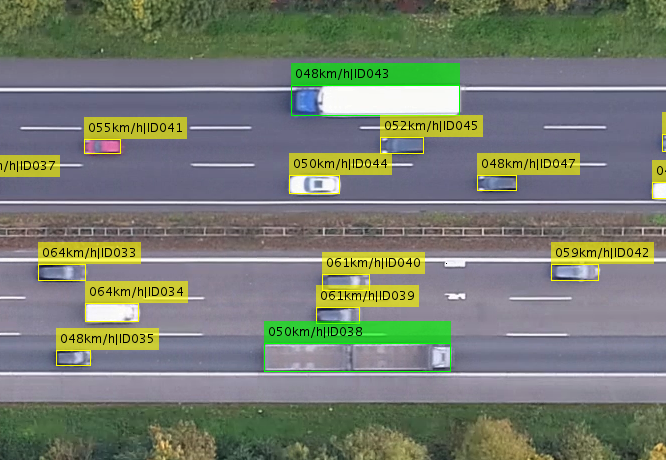}
		\caption{Example of vehicle positions that have been extracted from recordings of the highD dataset \cite{highDdataset}.}
		\label{fig:dronedetections}
\end{figure}

\subsection{highD Dataset}

The highD dataset \cite{highDdataset} consists of vehicle trajectories on German highways. Using a quadcopter, highway traffic has been recorded from a bird's eye perspective at six different sites around Cologne, Germany in 2017 and 2018. As a road segment of more than 400~m is covered, more than 20 vehicles in both directions are recorded on average at the same time (see Fig.~\ref{fig:dronedetections}). Finally, the road user behavior recorded with the drone videos is naturalistic. While measurement vehicles with e.g. 360° LiDAR on top could influence surrounding vehicles with their conspicuous sensors, the drone is not noticeable to drivers. The highD dataset includes 60 recordings with a total length of more than 16 hours on highways with two or three driving lanes per direction. The recordings contain about 110~000 vehicles split into about 90~000 cars and 20~000 trucks resulting in about 44~500 driven kilometers. The highD dataset \cite{highDdataset} is currently the largest publicly available naturalistic trajectory dataset of vehicles on highways.
However, the highD dataset does mainly contain rather higher speed scenarios, while the ALKS Regulation is defined for an operational speed of up to 60 km/h. Although the highD datasets provides a useful data basis for the proposed analysis, it would be more suitable to have a similar dataset containing lower speed scenarios caused by e.g. traffic jams or lower speed limits.

\section{Method}

\subsection{Overview}
\begin{figure}[h]
        \vspace*{0.2cm}
		\centering
		\includegraphics[width=0.48\textwidth]{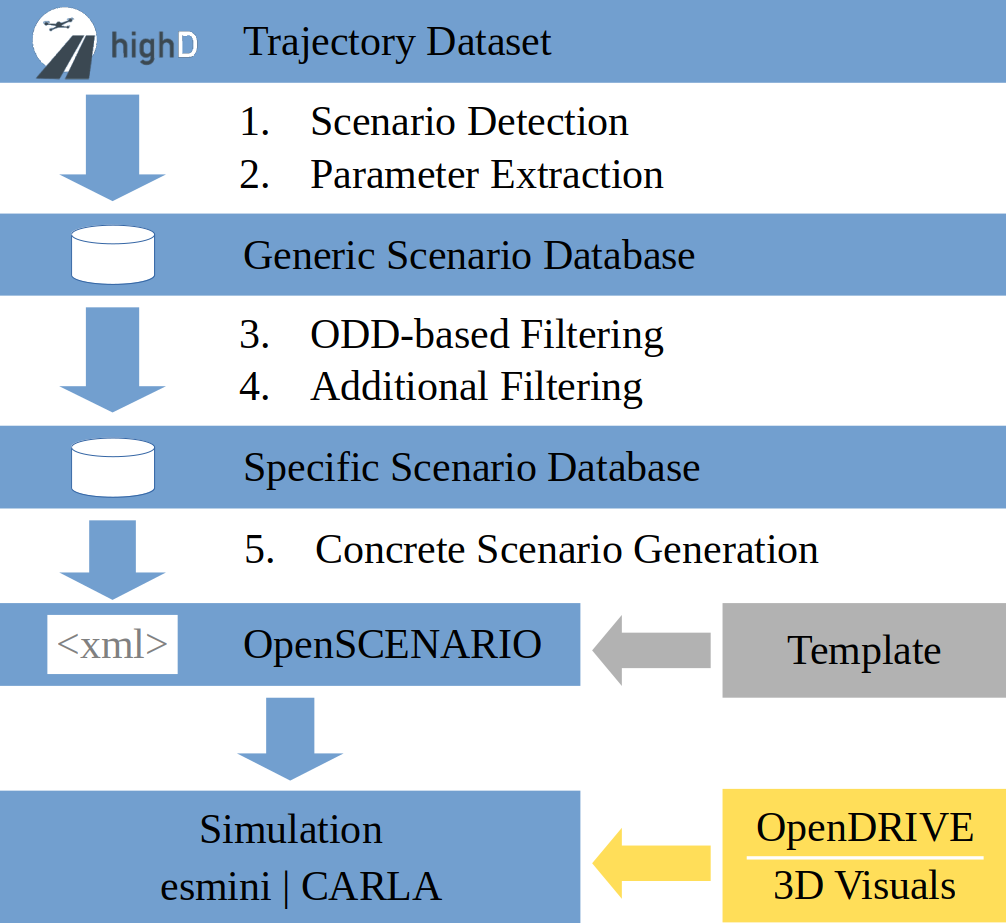}
		\caption{Methodology to generate concrete scenarios from highD for ALKS testing.}
		\label{fig:pipeline}
\end{figure}

We implemented the extraction scenarios from trajectory data as a multi-step process, see Fig.~\ref{fig:pipeline}. First, a generic database of scenarios is created, which is set up independently from any specific use case but instead aims for maximum comprehensiveness. The input data are the aforementioned recordings from the highD dataset, consisting of trajectories for every vehicle as well as lane assignments and relational information to surrounding road users. This time series data is analyzed from the point of view of every vehicle with respect to specific events like lane changes of the respective ego vehicle, lane changes of other (challenging) vehicles, brake maneuvers, etc. With the help of these events, various types of scenarios can be detected, including those defined in the ALKS Regulation. For each of these, there is a set of specific parameters describing the concrete scenario. Apart from the initial scene these also cover averages and extrema of the vehicles' motion states during the duration of the scenario. The parameter values are extracted and added to the database.

The subsequent step takes the Operational Design Domain (ODD) of the system under test into account, i.e. the database entries are filtered according to the system boundaries of ALKS systems as specified by the UNECE Regulation. Additional filter criteria can be applied to ensure relevance of the derived scenarios.

The parameters of the remaining scenarios are used to populate previously created scenario templates and generate individual OpenSCENARIO files. As the implementations of that standard by today's simulation tools are not (yet) fully compatible, separate output files are created for each supported tool. For now, this includes esmini and CARLA.

Finally, the generated OpenSCENARIO files can be used in simulation. There, apart from the OpenSCENARIO file two other building blocks are required: an OpenDRIVE file and a file containing the 3D visual information, like textures, posts, buildings and leafage.

The process for the creation of the OpenDRIVE map is based on an orthophoto of the recording location from the highD dataset. This orthophoto is used within the Mathworks RoadRunner \cite{roadrunner} software to create the OpenDRIVE file and the 3D visual database, both using the same point of origin as the highD dataset to get consistent coordinate systems.

\subsection{Scenario Selection}
There are six main categories of test scenarios specified in Annex 5, Chapter 4 of the ALKS Regulation. They have been split up into several sub-scenarios in order to satisfy the variations within each of the categories. Not all of them are equally well-suited to be derived from highD:
\begin{itemize}
    \item The field of view tests cover forward and lateral detection ranges regarding other road users and are not directly related to any specific driving scenario, thus do not benefit from real-world data. The same is true for free driving without immediate influence of surrounding traffic. 
    \item Other scenarios involve one or multiple stationary vehicles or road users blocking the lane. As highD focuses on regular traffic flow on German highways, these situations do not exist in any considerable quantity within the recorded trajectories.
    \item Crossing pedestrians and road user behavior inducing unavoidable collisions, i.e. actual accidents, are not contained at all in the dataset.
\end{itemize}
As a result of these considerations, the following scenarios were determined to benefit most from naturalistic real-world measurements:
\begin{itemize}
    \item "Lead Vehicle Brake": The ego vehicle drives at a constant speed, the preceding challenging vehicle in the same lane drives at a different or the same speed. At a trigger time point the latter decelerates abruptly.
    \item "Cut-In": The ego vehicle drives at a constant speed, the challenging vehicle in an adjacent lane drives at a different or the same speed. At a trigger time point the challenging vehicle changes into the ego vehicle's lane.
    \item "Swerving Lead Vehicle", "Swerving Side Vehicle": A vehicle in the same or an adjacent lane to the ego vehicle shows continuous oscillations in lateral position within the lane.
\end{itemize}
The two scenarios described in the last bullet point are of long-term nature and can be observed only partially by a stationary drone. As the brake and cut-in scenarios are event-oriented, they can be recorded in their entirety and are thus the focus of the published scenario set as well as this paper.

\subsection{Scenario Parameters}
The OpenSCENARIO standard offers the possibility to specify parameters for a scenario in a dedicated \emph{\textless ParameterDeclaration\textgreater}\space section at the beginning of the file \cite{asamOS}. This allows easy variation and separates them from the storyboard definition, which is constant within each type of scenario. For the method presented here, a file was created manually for each logical scenario and then used as a template for the generation of concrete scenarios by updating its parameters according to the database. The most relevant are listed in Table~\ref{table_parameters_brake}:

\begin{table}[h]
\caption{Parameters for scenario "Lead Vehicle Brake"}
\label{table_parameters_brake}
\begin{center}
\begin{tabular}{|l|c|}
\hline
Initial ego velocity & m/s\\
\hline
Initial challenging vehicle velocity & m/s\\
\hline
Initial distance to challenging vehicle & m\\
\hline
Challenging vehicle brake trigger distance & m\\
\hline
Challenging vehicle brake duration & s\\
\hline
Final challenging vehicle velocity & m/s\\
\hline
\end{tabular}
\end{center}
\end{table}

The brake scenario starts with the ego and the challenging vehicle driving at a defined distance in the same lane, but at different velocities. This initial scene is defined by the first three parameters. The duration of the deceleration is used as the defining quantity for the actual brake action. Also, while the UN Regulation mentions lead vehicle decelerations until standstill, this does not occur in highD recordings. Therefore, the list includes a parameter for the lead vehicle's final velocity at the end of the maneuver.
One general difference between the scenario definition used for the database and a file used as an input for a simulation tool is its starting time point. The former regards the start of the scenario with the initiation of the action defining it, i.e. the deceleration of the lead vehicle or the start of lateral motion for a vehicle cutting into the ego lane. The initial parameters are sampled at that time. In contrast, the corresponding actions in a simulation are usually supposed to happen after a short time span in order to give a system under test some initialization time. The initial distance from the database is therefore used as the trigger criteria for the maneuver, while the initial values exported to the OpenSCENARIO file are calculated so the trigger distance is reached at a simulation time of five seconds, under the assumption of constant velocities up to that point.

\begin{table}[h]
\caption{Parameters for scenario "Cut-In"}
\label{table_example}
\begin{center}
\begin{tabular}{|l|c|}
\hline
Initial ego velocity & m/s\\
\hline
Initial challenging vehicle velocity & m/s\\
\hline
Initial distance to challenging vehicle & m\\
\hline
Initial challenging vehicle relative lane & {-1,+1}\\
\hline
Initial challenging vehicle lane offset & m/s\\
\hline
Challenging vehicle cut-in trigger distance & m\\
\hline
Challenging vehicle cut-in distance & m\\
\hline
Final challenging vehicle velocity & m/s\\
\hline
Final challenging vehicle lane offset & m/s\\
\hline
\end{tabular}
\end{center}
\end{table}

As in the previous scenario, the definition of the cut-in scenario along with its list of parameters accounts for different initial velocities of the ego and the challenging vehicle as well as an initial distance. The sinusoidal lane-change itself is parameterized by the traveled distance during the lane change, which proved to generate trajectories matching the real-world measurements best, see Section~\ref{section:results}. As vehicles do not typically drive perfectly at the center of the lane, the challenging vehicle's offset in its initial as well as in its final lane can be specified. A parallel change of velocity of the challenging vehicle is also supported via the respective parameters. Again, the initial distance is calculated to trigger the actual maneuver at a simulation time of five seconds.

The parameters for the two "Swerving Vehicle" scenarios incorporate the variation range of the lateral offset within the lane and the maximum lateral acceleration of the lead or side vehicle.

Not included in either table are additional parameters for localizing the vehicles in the correct OpenDRIVE lanes with respect to the real-world observation.
Furthermore, highD data includes the dimensions of all vehicles with respect to length and width as well as their classification like car, bus or truck. As a consequence, the vehicles referenced in the generated concrete scenarios are not taken from standard vehicle catalog definitions but instead match their features from the respective real-world measurements.

\subsection{Scenario Extraction Conditions}
In order to generate test cases relevant for ALKS verification, the specified system boundaries in the UNECE Regulation provide a way to narrow down the number of scenarios. While the highD dataset covers German highways and thus matches the ALKS requirements regarding road and infrastructure very well, the vast majority of vehicles travel at typical velocities of 100 to 130~km/h, exceeding the ALKS limit of 60 km/h. To remove scenarios deviating too much from that limit, the database has been filtered using a threshold of 70 km/h for the initial velocity of the ego vehicle. For the lead vehicle brake scenario, an additional filter condition is applied: to differentiate deliberate braking from regular velocity adjustments, only scenarios in which the lead vehicle exceeds a certain deceleration threshold are considered as relevant. Here, this value has been chosen as 2~m/s² \cite{haas2004use}. A swerving vehicle scenario is detected if a vehicle's lateral position shows a variation range of at least 1.2 m within the measurement area. The resulting numbers of all scenarios are displayed in Fig. \ref{fig:scenario_numbers}:

\begin{figure}[h]
        \vspace*{0.2cm}
		\centering
		\includegraphics[width=0.48\textwidth]{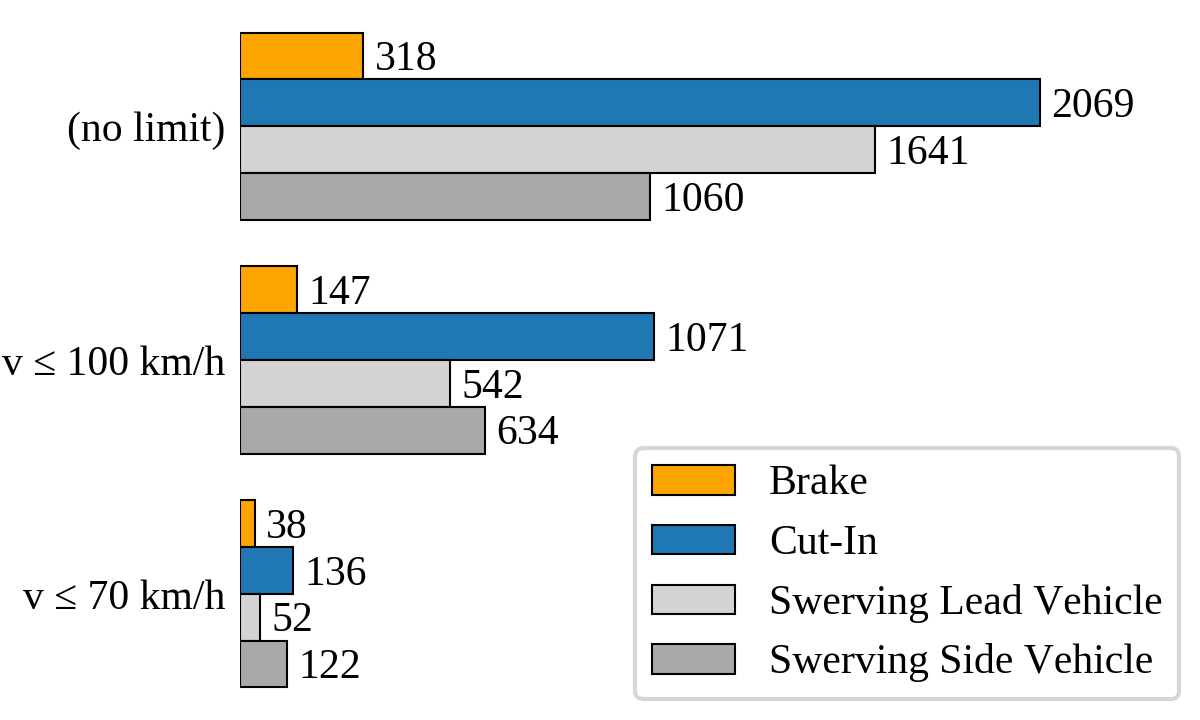}
		\caption{Number of scenarios in total and after the application of various velocity thresholds.}
		\label{fig:scenario_numbers}
\end{figure}

Another condition for event-centric scenarios to be extracted was their completion within the measurement area. Cut-in and brake scenarios whose start and/or end is not visible have been discarded. This was necessary to exclude canceled or double lane-changes that would have skewed the data, and to reliably determine the initial and final velocity of the lead vehicle.

\subsection{Simulation of Scenarios}

As mentioned above, esmini and CARLA were picked for the exemplary replay of the generated scenarios in simulation. Running the extracted scenarios in esmini only requires the parameterized OpenSCENARIO file, the OpenDRIVE file and the OpenSceneGraph file. In contrast to esmini, CARLA version 0.9.10 does not support all OpenSCENARIO attributes that were used \cite{carlaOSC}. Therefore, the files had to be adapted as follows.

For CARLA a role type had to be added to each actor and an environment description specifying weather and road conditions was inserted.

One of the main differences between esmini and CARLA is the inverse interpretation of the relative lane ID, which is used both during the initialization of the challenging vehicle and the lane change. 
Additionally, CARLA only supports the assignment of a target lane relative to the actor itself and the sinusoidal lane change had to be alternated to linear. For improved usability of the scenarios in CARLA, a set of CARLA specific triggers, who e.g. check for wrong lanes or collisions, and allow to test a function and evaluate the scenarios, were added to the templates.

Finally, as the the generated scenarios are intended to optimize and validate different Automated Lane-Keeping Systems the OpenSCENARIO files contain placeholders for the definition and parameterization of actor controllers. As defined in OpenSCENARIO, these controllers are intended to specify how vehicles or pedestrians should be controlled and define the specific behavior of smart actors \cite{asamOS}. 

After the parameterization of the OpenSCENARIO files, XML checker was applied and the scenarios were replayed in the two simulation environments. As an example, Fig.~\ref{fig:deceleration_dist_max} shows the simulation of a cut-in scenario in CARLA OpenDRIVE standalone mode and esmini. 
\begin{figure}[h]
        \vspace*{0.2cm}
		\centering
		\includegraphics[width=0.48\textwidth]{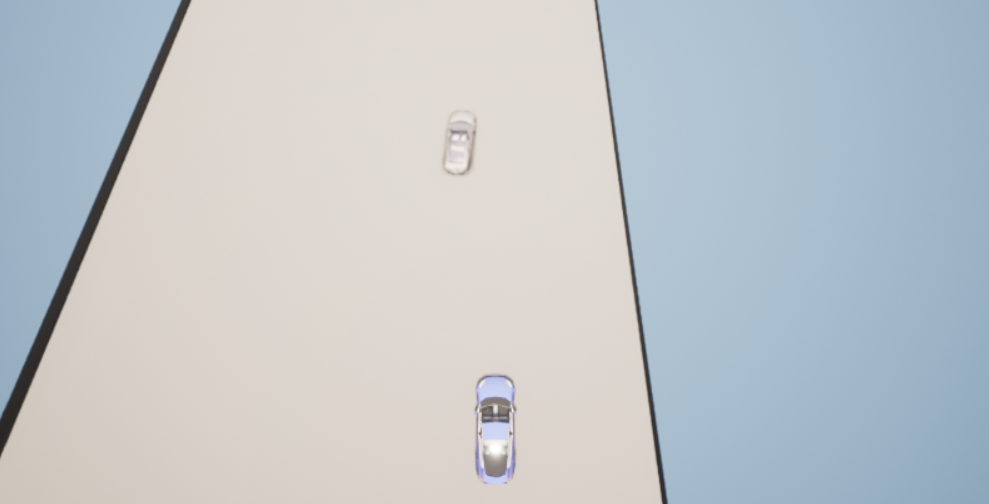}
		\includegraphics[width=0.48\textwidth]{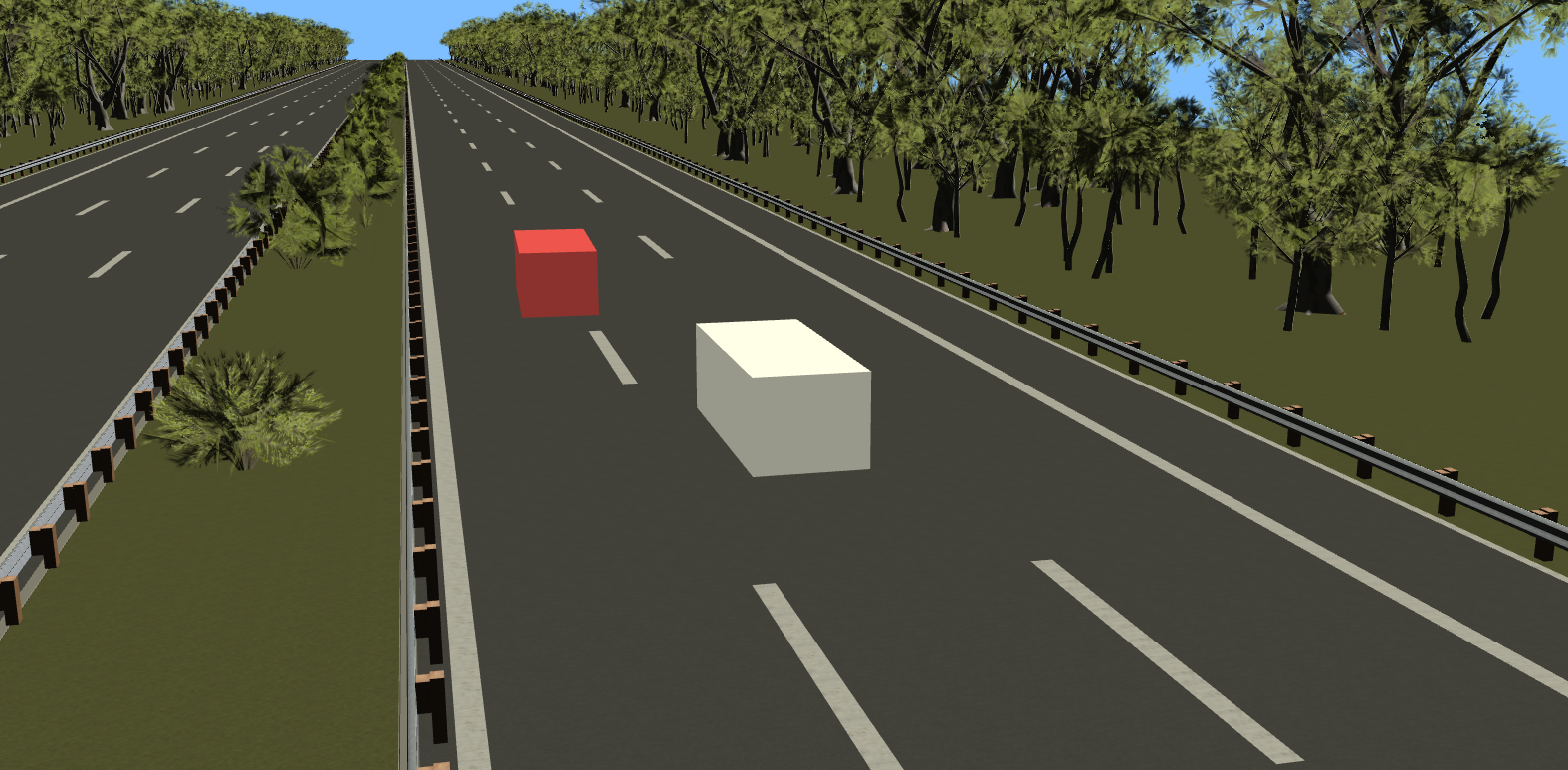}
		\caption{Simulation of a cut-in scenario with CARLA [upper image] in OpenDRIVE standalone mode and esmini [lower image].}
		\label{fig:sim_screenshot}
\end{figure}

\section{Results}
\label{section:results}
\begin{figure}[h]
        \vspace*{0.2cm}
		\centering
		\includegraphics[width=0.48\textwidth]{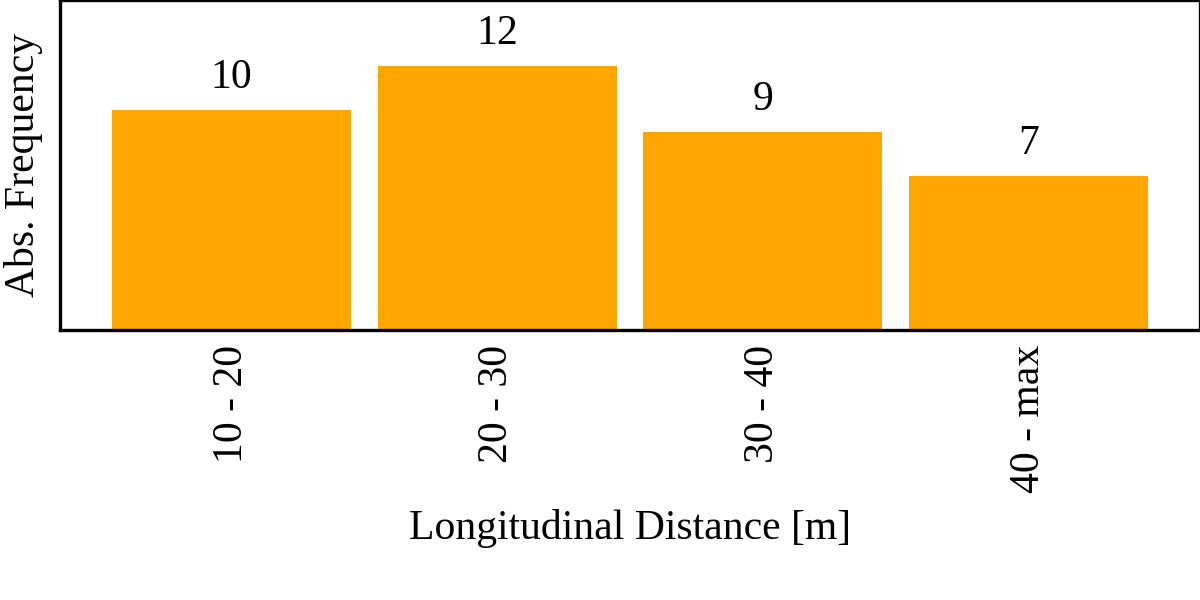}
		\includegraphics[width=0.48\textwidth]{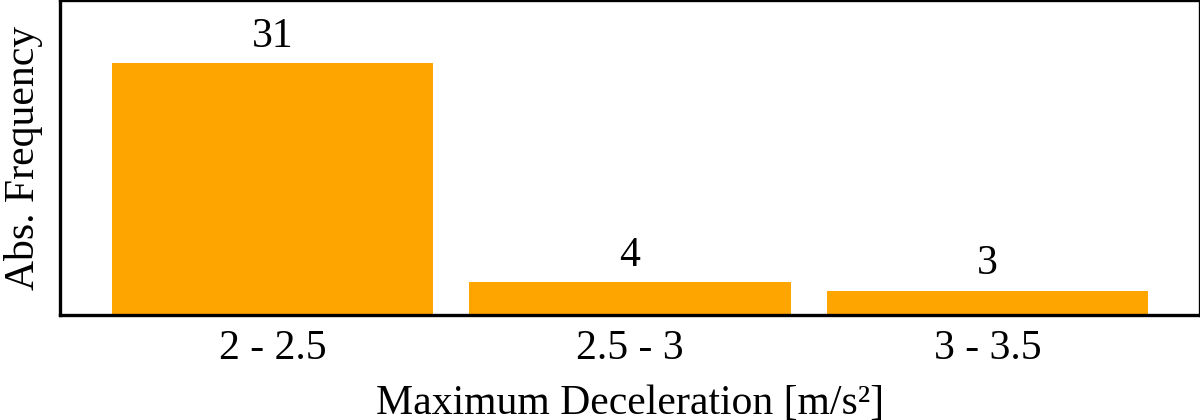}
		\caption{Histogram of longitudinal distance and maximum deceleration in the "Lead Vehicle Brake" scenario.}
		\label{fig:deceleration_dist_max}
\end{figure}

After filtering the scenario database according to ALKS system boundaries as well as criticality requirements, 136 cut-in scenarios and 38 brake scenarios are ultimately extracted. For the former, the distributions of the longitudinal distance between the ego vehicle and the lead vehicle as well the maximum deceleration of the lead vehicle are shown in Fig.~\ref{fig:deceleration_dist_max}. 

Only in very few instances the deceleration surpasses 2.5~m/s². As the greatest deceleration is 3.3~m/s², the value of 6~m/s² specified in the Regulation (Annex~5, Chapter~4.3 (f)) is not reached at all. These magnitudes are required for collision avoidance situations. Even though the highD data contains deceleration scenarios with a deceleration greater than 6~m/s², the velocity range of 70~km/h excludes a significant quantity of such highD recordings.

\begin{figure}[h]
        \vspace*{0.2cm}
		\centering
		\includegraphics[width=0.48\textwidth]{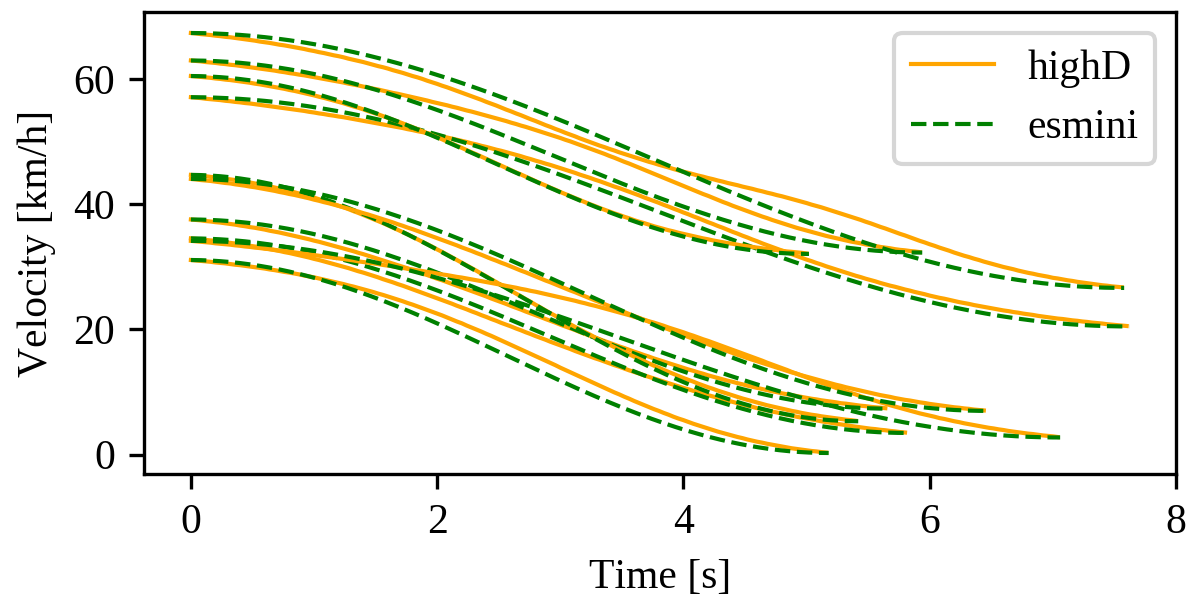}
		\caption{Challenging vehicle velocity in esmini simulation and highD during the deceleration phase for ten concrete brake scenarios.}
		\label{fig:esmini_brake}
\end{figure}

Fig.~\ref{fig:esmini_brake} displays a comparison between the change of velocity of the real-world and the simulated challenging vehicle for a subset of ten randomly chosen concrete scenarios. The tool of choice was esmini, as its simplistic physics and controller implementation results in an exact realization of the specified geometric primitives, thus allowing a study of the scenario description itself without influence of controller quality. The velocity change of the lead vehicle within the parametrized duration was modeled using a cubic function, which results in a root mean square error (RMSE) of 4.49~km/h for the difference in velocity between the source data and esmini simulation as an average for all brake scenarios. This value is driven up by individual scenarios where the deceleration is applied stepwise or in comparable ways that can not be described adequately in OpenSCENARIO.

\begin{figure}[h]
        \vspace*{0.2cm}
		\centering
		\includegraphics[width=0.48\textwidth]{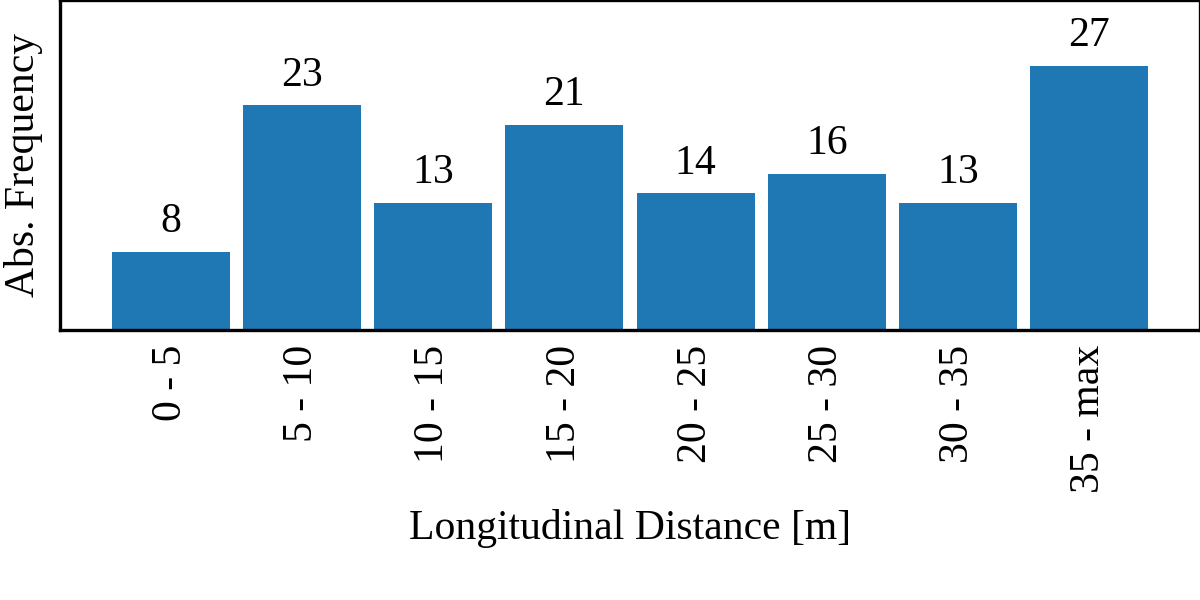}
		\includegraphics[width=0.48\textwidth]{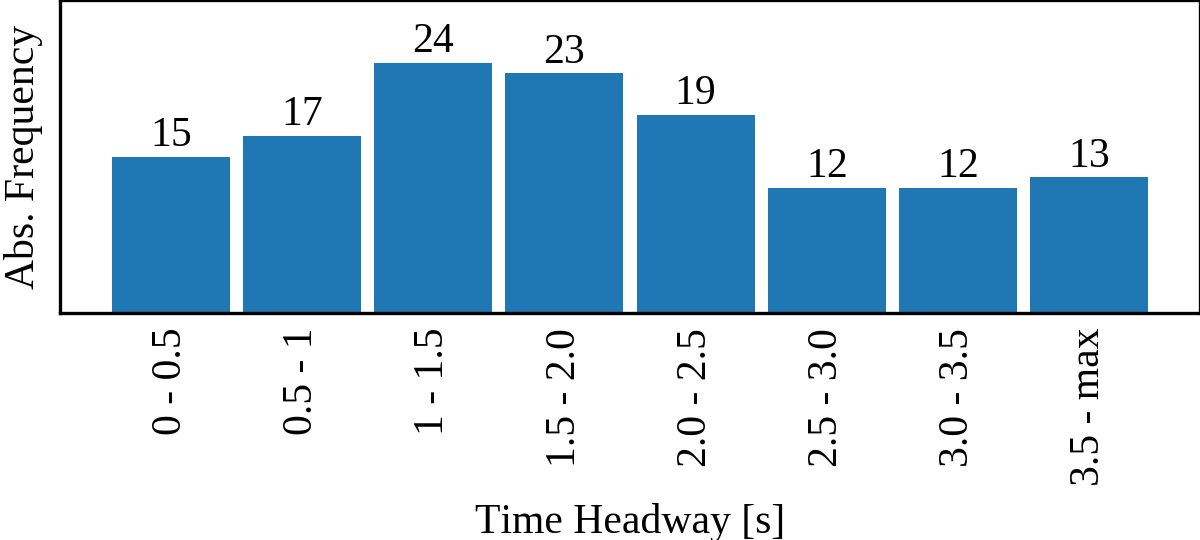}
		\caption{Histogram of longitudinal distance and time headway at maneuver start in the "Cut-In" scenario.}
		\label{fig:cutin_dist_thw}
\end{figure}

\begin{figure}[h]
        \vspace*{0.2cm}
		\centering
		\includegraphics[width=0.48\textwidth]{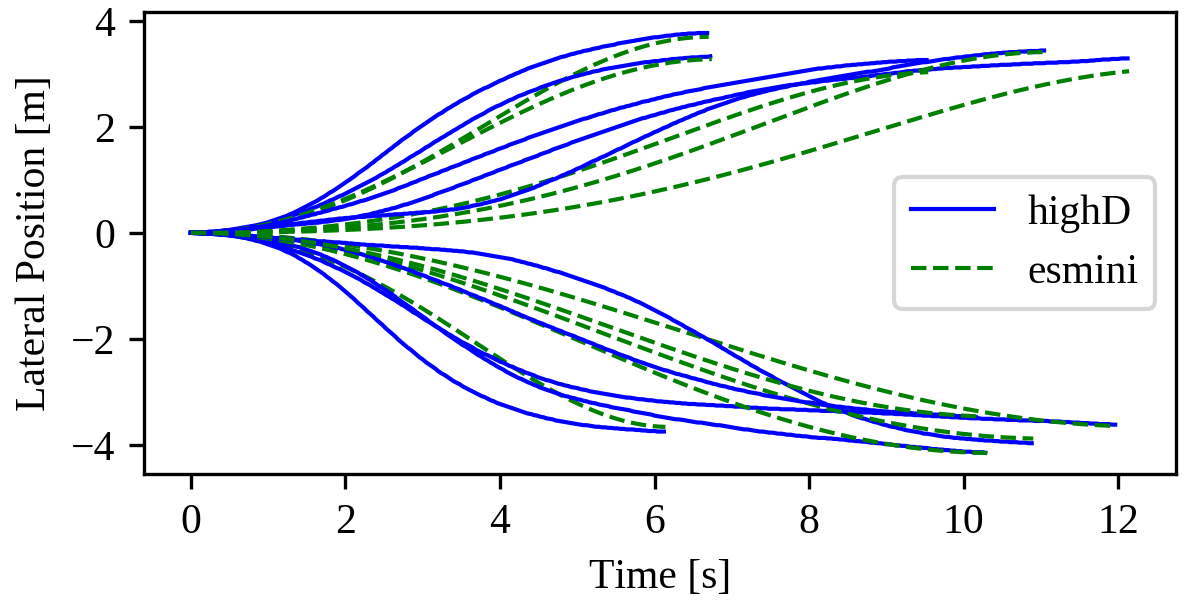}
		\caption{Lateral position of the challenging vehicle in esmini simulation and highD for ten concrete cut-in scenarios.}
		\label{fig:esmini_cutin}
\end{figure}

For the cut-in scenario, distributions of longitudinal distance and time headway (THW) at the start of the cut-in maneuver are shown in Fig.~\ref{fig:cutin_dist_thw}. As expected for the limited range of velocities, the distances include close ranges and even eight cut-ins starting at under 5~m distance. The peak of the THW distribution is found between one and two seconds.
Fig.~\ref{fig:esmini_cutin} shows ten exemplary lateral motion trajectories for the real-world challenging vehicles and their simulated counterparts. The parameterization of the sinusoidal curve using the length of the maneuver guarantees matching end points. In between, the simulated symmetric curves do differ from their source data and tend to result in slightly less aggressive lane changes. The RMSE for the deviation of lateral position is 0.40~m  as an average for all cut-in scenarios. Better fitting would require the availability of more complex shapes with more parameters.

\section{Conclusion}

To overcome the need of a reasonable parameterization that is required for the test scenarios of the ALKS Regulation, this paper highlights a methodology to derive meaningful parameters based on real-world data. The results proof the methodology's capability to generate representative scenarios in the simulation describing similar vehicle trajectories. The use of real-world data underlines the validity of the scenarios due to considering the heterogeneous behavior of the relevant traffic participants. The derived scenarios are published in OpenSCENARIO and OpenDRIVE to be replayed in frequently used simulation tools like CARLA and esmini. These scenarios can be useful for the development of driving functions, but also for testing an implemented ALKS function in simulations. The dataset is available at https://www.levelXdata.com/scenarios. For the future, we plan to analyze additional real-world data to increase the validity. We will also conduct analyses to perform a statistically representative survey of the ALKS parameters in order to be able to prove a saturation of the data. The goal is to ensure that extreme values, such as the deceleration in the braking scenario, but also the general distribution of other parameters like weather are taken into account. In addition, it is foreseeable that complexity of scenarios needs to be increased e.g. by adding features like occlusions and action constraints.

\bibliographystyle{IEEEtran}
\bibliography{main}


\end{document}